\documentclass{article}

\usepackage[preprint]{neurips_2024}
\usepackage[utf8]{inputenc} 
\usepackage[T1]{fontenc}    
\usepackage{hyperref}       
\usepackage{url}            
\usepackage{booktabs}       
\usepackage{amsfonts}       
\usepackage{nicefrac}       
\usepackage{microtype}      
\usepackage{xcolor}         
\usepackage{graphicx}
\usepackage{longtable}
\usepackage{multirow}
\usepackage{amsmath}

\usepackage{listings}
\usepackage{xcolor,colortbl}
\definecolor{darkgreen}{rgb}{0.0, 0.5, 0.0}

\title{Fine-Tuning In-House Large Language Models to Infer Differential Diagnosis from Radiology Reports}

\author{
  Luoyao Chen \\
  NYU Grossman School of Medicine \\
  New York, NY 10016 \\
  \texttt{Luoyao.Chen@nyulangone.org} \\
  \And
  Revant Teotia \\
  New York University\\
  New York, NY 10012 \\
  \texttt{rt2741@nyu.edu} \\
  \AND
  Antonio Verdone \\
  NYU Grossman School of Medicine \\
  New York, NY 10016 \\
  \texttt{Antonio.Verdone@nyulangone.org} \\
  \And
  Aidan Cardall \\
  NYU Grossman School of Medicine \\
  New York, NY 10016 \\
  \texttt{Aidan.Cardall@nyulangone.org} \\
  \And
  Lakshay Tyagi \\
  New York University\\
  New York, NY 10012 \\
  \texttt{lt2504@nyu.edu} \\
  \And
  Yiqiu Shen \\
  NYU Grossman School of Medicine \\
  New York, NY 10016 \\
  \texttt{Yiqiu.Shen@nyulangone.org} \\
  \And
  Sumit Chopra \\
  NYU Grossman School of Medicine \\
  New York, NY 10016 \\
  \texttt{Sumit.Chopra@nyulangone.org} \\
}

\begin{document}

\maketitle

\begin{abstract}
   Radiology reports summarize key findings and differential diagnoses derived from medical imaging examinations. The extraction of differential diagnoses is crucial for downstream tasks, including patient management and treatment planning. However, the unstructured nature of these reports, characterized by diverse linguistic styles and inconsistent formatting, presents significant challenges. Although proprietary large language models (LLMs) such as GPT-4 can effectively retrieve clinical information, their use is limited in practice by high costs and concerns over the privacy of protected health information (PHI). This study introduces a pipeline for developing in-house LLMs tailored to identify differential diagnoses from radiology reports. We first utilize GPT-4 to create 31,056 labeled reports, then fine-tune open source LLM using this dataset. Evaluated on a set of 1,067 reports annotated by clinicians, the proposed model achieves an average F1 score of 92.1\%, which is on par with GPT-4 (90.8\%). Through this study, we provide a methodology for constructing in-house LLMs that: match the performance of GPT, reduce dependence on expensive proprietary models, and enhance the privacy and security of PHI.
   
\end{abstract}

\section{Introduction}
Radiology reports document the findings from medical imaging studies (~\cite{reiner2007radiology}). A typical radiology report includes several key components: the patient's prior medical history, the imaging technique employed,  the radiological findings (observed abnormalities or notable features in the images), and the \textit{impression} (the radiologist’s interpretation of these findings and suggests a ranked list of possible medical conditions).  Differential diagnosis, which can often derived from the \textit{impression}, outlines potential conditions that could account for the abnormalities observed, guiding subsequent clinical decisions.

Automatically determining the differential diagnosis from radiology reports is a crucial task  (\cite{farmer2020enhancing}). First, such a system could enhance clinical decision support by integrating diagnostic suggestions directly into clinical systems, which aids in reducing errors and improving patient outcomes. Second, this system also supports research and training by providing a rich dataset for developing diagnostic tools and educational resources for trainee physicians. Third, systematic documentation facilitated by automation ensures compliance with healthcare regulations and improves the quality and accountability of medical practices.

The introduction of Large Language Models (LLMs) provides a promising avenue for automating the determination of differential diagnoses in medical contexts. Commercial LLMs such as GPT-4 (~\cite{GPT}) have demonstrated effectiveness in various clinical information retrieval tasks; however, they pose several challenges. First, the substantial costs associated with deploying these models on the vast number of daily radiology reports can be prohibitive (\cite{brady2024developing}). Second, despite being trained on extensive textual data, these proprietary models often lack the specialized expertise required for accurately interpreting radiology reports, particularly in distinguishing between similar or related symptoms. Lastly, concerns regarding the confidentiality of patient health information (PHI) and compliance with healthcare regulations have eroded public trust in the use of commercial LLMs within healthcare settings  (\cite{ullah2024challenges}). Because of these issues, developing in-house LLMs tailored to the specific needs is a more viable solution. This approach not only mitigates concerns about PHI security but also ensures that the models are fine-tuned to address the unique challenges of medical diagnostics (\cite{chen2024burextract}).

To train such in-house LLMs, a substantial amount of annotated data is required, wherein radiologists meticulously mark the pathologies associated with each radiology report. In this work, we demonstrate that open-source LLMs can be trained to match the effectiveness of GPT-4 without requiring human-annotated labels, using radiology musculoskeletal (MSK) reports.

\section{Related work}
Information Extraction is a well-established application of NLP to pathology. It encompasses a range of tasks, including Named Entity Recognition (NER), report classification, and key concepts extraction (\cite{lopez2022natural}). Methods have matured from basic rule-based to traditional machine learning and transformer-based approaches, and more recently, to generative-based approaches.

\paragraph{Basic attempts} use rule-based methods that usually utilize a set of pre-defined patterns for string matching, which is sensitive to minor changes and is less flexible to negations  (\cite{zeng2023improving}). 

\paragraph{Traditional machine learning methods} Before the prevalence of Transformer (\cite{vaswani2017attention}), traditional machine learning methods enabled more diverse analysis. For example, CRF (\cite{patrick2009cascade}), SVM (\cite{tang2013recognizing}), and XGBoost (\cite{ryu2020prediction}) are common methods for NER or classification. To better encode the context information resulting from word order, Peters et.al uses Bi-direction RNN  (\cite{peters2017semi}), and Wang et al. use LSTM (\cite{wang2019multitask}). Furthermore, to break the limit of word ordering and variations of the same expression, Yoon et.al apply Text GCN, which uses a graph of the word as text representation, with edges determined by the word co-occurrence (\cite{yoon2019information}).

\paragraph{Transformer-based models} With the advent of the Transformer, BERT (\cite{zeng2023improving}) and RoBERTa (\cite{liu2019roberta}) become widely used, including subsequent models such as BioBERT (\cite{lee2020biobert}) that pre-train BERT using biomedical text. The self-attention mechanism in the Transformer uses different positions of a single sequence to compute a representation of the sequence on a global scale (\cite{lopez2022natural}), which leads to its superior performance in discriminative classification compared with traditional machine learning models. 

\paragraph{Generative models} Recently, Large Language Models, from commercial models GPT-4 (\cite{GPT}) to open-sourced models Llama-3 (\cite{Llama3modelcard}), Mistral  (\cite{mistralv3modelcard}), and domain-fine-tunedd LLM such as MediTron (\cite{chen2023meditron}), have fundamentally changed the field of natural language processing. The generative capability removes the restriction of pre-defined classes. By pretraining on a large volume of text, LLM acquires knowledge about both language and domain and is capable of performing a variety of tasks, from summarization to question answering.

\begin{figure}[!t]
    \centering
    \includegraphics[width=\linewidth, height=6cm]{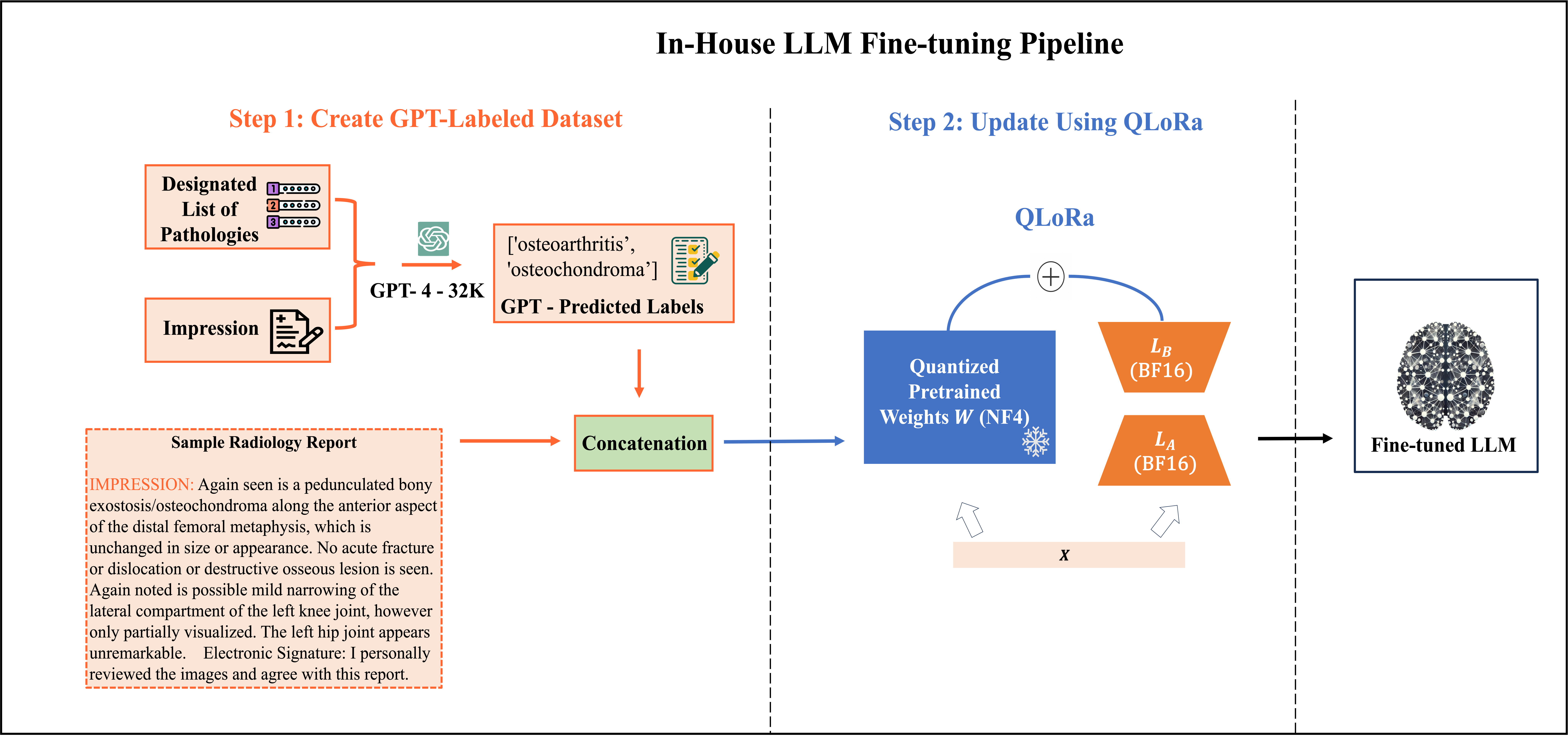}
    \caption{Pipeline for fine-tuning in-house LLM}
    \label{fig:pipeline}
\end{figure}

\section{Methods}
\subsection{Problem formulation}
Let $I$ denote the impression section of a radiology report. Our goal is to build a model that maps $I$ to a list of differential diagnoses: $I \rightarrow [p1, p2…]$, as illustrated in Figure\ref{fig:pipeline}. We list all 133 possible differential diagnoses in Appendix \ref{sec:appendixB}.

\subsection{In-house LLM fine-tuning pipeline}

\paragraph{Step1: generate training labels using GPT-4-32K} To fine-tune the in-house LLMs, we need annotated training data in which each impression is associated with a list of differential diagnoses. To obtain training labels without requiring human annotation, we use GPT-4-32K to identify the diagnosis mentioned in the impression. For this, we devise an efficient prompt—detailed in Appendix \ref{sec:appendixA.1}. For each report, we run GPT-4 three times with a temperature of 1.0, and the most common output from these runs is selected via majority voting. As shown in the example in Figure \ref{fig:pipeline} step 1, the output for this step consists of an impression labeled with a list of two pathologies, termed GPT-predicted labels. We divide this dataset into 18,538 reports for training and 12,518 for validation. We apply stratified iterative sampling to maintain per-class even distribution between training and validation  (\cite{pmlr-v74-szymański17a, tang2013recognizing}).

The validity of GPT-generated labels is validated later by comparing them against human-annotated labels on a small test set of 1,067 reports. As illustrated in Table\ref{tab:result}, GPT-4-32k achieves an average  F1 Micro score of 90.8\%, which demonstrates the reliability of these training labels.

\paragraph{Step2: fine-tune in-house LLMs using QLoRa}
We fine-tune Llama-3-8B, Llama-3-8B-instruct (\cite{Llama3modelcard}), Mistral-v0.3, Mistral-v0.3-instruct (\cite{mistralv3modelcard}) using QLoRa (\cite{dettmers2024qlora}), which has a 4-bit NormalFloat storage type and a 16-bit BrainFloat (\cite{kalamkar2019study}) computational type. To reduce memory usage and computational costs, model weights are quantized to a lower precision (4-bit). During fine-tuning, to prevent precision loss, the quantized weights are dequantized back to higher precision (16-bit BrainFloat) before being used to update the LoRa adapter. Alongside dequantization, QLoRa incorporates LoRa (\cite{hu2021lora}), which consists of a list of trainable adapters, with the latter being a product of low-rank matrices that store critical information about the task. Specifically, for each linear module in the attention head, whose pre-trained weight denoted $W_{d\times k}$, we approximate the weight using a LoRa adapter, consisting of two add-on low-rank matrices, $L_A({d\times r})$ $\times$ $L_B({r\times k})$, where r is the rank of each adapter. Together, each LoRa adapter is updated by
$$W \leftarrow W+\Delta w, \space\space \Delta w=\frac{\alpha}{r}L_BL_A$$
Where W is the pre-trained weight, alpha acts like a learning rate, and r is the rank. We use autoregressive CrossEntropy (CE) on the generated tokens during fine-tuning, i.e. exclude the loss contributed by user prompt. This process is shown in Figure \ref{fig:pipeline}, step 2.

After using QLoRa, the size of the trainable parameter has significantly reduced to less than 5\%, as indicated in Table \ref{tab:ParamSize}. The fine-tuning prompt can be found in Appendix \ref{sec:appendixA.2}.

\begin{table}[hbt!]
  \centering
  \small
  \begin{tabular}{ccc}
    \hline
    \textbf{Model} & \textbf{Original Size} & \textbf{LoRa Trainable} \\
    \hline
    Llama-3-8b/instruct     & {4,540,600,320 (16GB)}   & {167,772,160 (40MB) (3.56\%)}   \\
    Mistral-v0.3/instruct     & {3,758,362,624 (14GB)}           & {167,772,160 (40MB) (4.27\%)}     \\
    \hline
  \end{tabular}
    \caption{Trainable Parameters Using QloRa}
    \label{tab:ParamSize}
\end{table}

\section{Experiments and results}

\subsection{Dataset}
Our dataset comprises 31,056 radiology reports from the musculoskeletal imaging department of a large academic medical center, spanning various imaging modalities (MRI, ultrasound, CT, X-ray) and anatomical regions (knee, wrist, pelvis, shoulder, spine, etc.). We cataloged a list of 133 differential diagnoses (see Appendix \ref{sec:appendixB}) representing the majority of common pathologies in musculoskeletal imaging. For the hold-out test dataset, an additional 1,067 impressions were collected and the labels for each were refined by radiologists. This refined dataset, designated as the gold standard test set, addresses potential inaccuracies from GPT-generated labels and provides a reliable metric for evaluation.


\subsection{Experiment setup}

We fine-tuned BERT, Llama-3-8b, Llama-3-8b-instruct, Mistral-7b-v.03, Mistral-7b-instruct-v.03. For LLMs, we used the QLoRa with rank 64, dropout 0.05, bias none, alpha 16 and linear target module.  For each LLM, we employed the AdamW\_8bit optimizer with a learning rate of 3e-4 and a weight decay of 0.01, alongside a learning rate scheduler that reduces on a plateau. Models were trained for 5 epochs. For the Llama-3 models, we used a batch size of 128, while Mistral models were trained with a batch size of 192. Training was conducted on four A100 (80GB) GPUs, with durations ranging from 8.5 hours for the base model to 9 hours for the instruct model. Results were averaged across three different seeds to ensure reliability.

\subsection{Evaluation metrics}
Given that each impression can receive multiple labeled diagnoses, we employed four metrics to evaluate the proposed pipeline: Precision, Recall, F1 Micro, and F1 Macro. For generative models, \emph{precision} decreases when the model exhibits hallucination (i.e. none-stop or incorrect outputs). \emph{Recall} measures the model's ability to correctly generate all relevant outputs. Furthermore, since our test set is imbalanced across 133 classes, we report both \emph{micro F1} and \emph{macro F1}, to provide a holistic view of performance across instances and classes, respectively.

\subsection{Results and discussions}

\begin{figure}[!t]
    \centering
    \includegraphics[width=\linewidth, height=6cm]{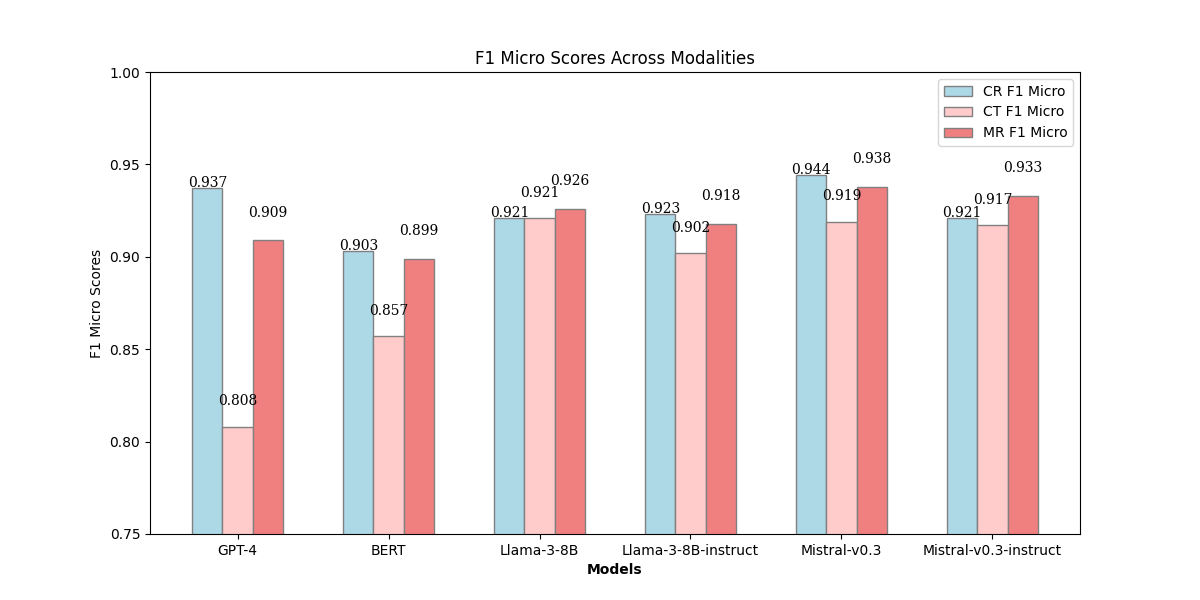}
    \caption{F1-micro across different modalities}
    \label{fig:modality}
\end{figure}

\begin{table*}[!ht]
\centering
\normalsize
\begin{tabular}{cccccc}
\hline
& \multicolumn{4}{c}{MSK30K} \\ 

& \multicolumn{1}{c}{Model} & \multicolumn{1}{c}{Precision} & \multicolumn{1}{c}{Recall} & \multicolumn{1}{c}{F1 Micro} & \multicolumn{1}{c}{F1 Macro}  \\ 

\hline

\multirow{4}{*}{W/O FT} & \multicolumn{1}{c}{GPT-4} & 0.941 & 0.877 & 0.908 & 0.868  \\ 
                         & \multicolumn{1}{c}{Llama-3-8b} & 0.097 & 0.022 & 0.036 & 0.031  \\ 
                        & \multicolumn{1}{c}{Llama-3-8b-instruct} & 0.770 & 0.473 & 0.586 & 0.563  \\ 
                        & \multicolumn{1}{c}{Llama-3-70b-instruct} & 0.944 & 0.585 & 0.722 & 0.681  \\ 

\hline
\multirow{4}{*}{FT} & \multicolumn{1}{c}{BERT} & 0.925$\pm$0.008 & 0.870$\pm$0.003 & 0.897$\pm$0.002 &	0.835$\pm$0.004 \\ 
                            
                            & \multicolumn{1}{c}{Llama-3-8b} & 0.932$\pm$0.013 & 0.907$\pm$0.011 & 0.919$\pm$0.004 & 0.904$\pm$0.005 \\ 
                            
                            & \multicolumn{1}{c}{Llama-3-8b-instruct} & 0.929$\pm$0.003 & \textbf{0.912}$\pm$0.004 &	\textbf{0.920}$\pm$0.002 &	\textbf{0.907}$\pm$0.006 \\ 
                            
                            & \multicolumn{1}{c}{Mistral-v.03} & 0.933$\pm$0.011& \textbf{0.919}$\pm$0.015 & \textbf{0.926}$\pm$0.011 & \textbf{0.913}$\pm$0.008 \\ 
                            
                            & \multicolumn{1}{c}{Mistral-v.03-instruct} & 0.938$\pm$0.009 & 0.910$\pm$0.005 & 0.918$\pm$0.012&	0.902$\pm$0.011  \\

\hline

\end{tabular}
\label{tab:result}
\caption{Performance of baseline and fine-tuned models. All models are trained using three different random seeds.}
\end{table*}

We evaluate the efficacy of our pathology extraction pipeline by comparing it with GPT-4’s inference scores on the test datasets, with the latter obtained by Appendix \ref{sec:appendixA.1}. Table~\ref{tab:result} showcases the results. In addition, we evaluate the performance across different modalities (CR, CT, MRI), which shows the effectiveness of our pipeline across modalities (Figure~\ref{fig:modality}). We summarize our results in Table \ref{tab:result} as below:




\begin{enumerate}

 \item Without fine-tuning (WO FT), open source models (Llama and Mistral) significantly fall behind commercial models such as GPT-4 (F1 micro 72\% versus 90.77\%). Such a gap highlights the need for fine-tuning LLMs to make them suitable for radiology impression pathology extraction.
 
 \item After fine-tuning, all LLMs exceed GPT-4 in Recall (exceed by 4.2\%), Micro F1 (exceed by 1.8\%) and Macro F1 (exceed by 4.5\%), and have on-par performances with GPT-4 for Precision (within 0.7\% difference). This indicates that the fine-tuned models are well able to capture the pathologies present in the impression and achieve satisfactory average performances across all classes. 
 
 \item Comparing generative (LLM) with discriminative models (BERT), all fine-tuned LLM significantly exceed the fine-tuned BERT in Recall, Micro F1, and Macro F1. This indicates that the fine-tuned generative model is more capable of understanding the complex context compared with the discriminative baselines. 

 \item The fine-tuned Llama-3-8b-instruct and Mistral model provide the highest Micro F1 (92.6\%). This indicates the success of the pipeline, showing that a 1000 times smaller open source model (8B parameters) can be trained to exceed the performance of large commercial LLMs (1.8T parameters). 

\end{enumerate}


\paragraph{Error analysis} We conduct an error analysis to analyze where the in-house LLMs fail by analyzing the confusion matrices for both the Llama-3-8b-instruct and Mistral-v0.3 models. Although these models have had high precision and recall, there are a small handful of pathologies that are still mispredicted. For example, a commonly confused pair is mis-predicting \textbf{partial rotator cuff tear} as \textbf{biceps tear}. This can be explained by the fact that the two pathologies share multiple similar components, anatomically and in their imaging and text representations. Namely, both pathologies occur at proximity anatomical locations (shoulder), share numerous common symptoms (pain in the front of shoulder, weakness when lifting and rotating arm, and difficulty in shoulder movement), and show similar radio-graph features(pattern of tendon damage). Consequently, their text representations will be similar (common keywords), leading to mis-predictions. However, note that such confusions are rare, i.e. less than 5 pathology pairs out of all possible pairs of 133 pathologies.

\section{Conclusion}
In this work, we provide a pipeline for constructing in-house LLMs without the need for human-annotated labels. This pipeline, once deployed, can facilitate automatic differential diagnosis extraction, and as a result, can increase diagnosis accuracy. Note that, as the proposed pipeline operates on radiology impressions written in natural language text, it can be generalized to other sections (Neuroradiology, Oncologic Imaging, Pediatric Radiology) as well, and which be in our future work. 

In sum, by achieving on-par or exceeding performance than GPT-4 on MSK dataset with only 5\% trainable parameters of open-sourced LLMs, we have proved the efficacy and potential for this method to be deployed in-house, as a replacement for proprietary GPT. As such, we will reduce costs, and ensure safety, ethnicity, and policy compliance.

\bibliography{ref.bib}

\appendix
\section{Appendix A: Prompts}
\subsection{GPT-4-32k prompt}\label{sec:appendixA.1}
\begin{quote}
    "You are a musculoskeletal radiologist reading the radiology impression below. Your task is to list only those pathologic conditions from the list of pathologies below that are explicitly mentioned in the radiology impression as either possible or definite. The pathology should be included strictly if it is specifically named in the report. Do not infer the presence or absence of a pathology that is not explicitly named in the report. If a pathology is not clearly mentioned, it should not be included in the output. If a pathology is specifically excluded with phrases such as "no fracture", or "without evidence of", or "no evidence of", or "without", or "within normal limits", please list that pathology as ABSENT. Present your answer in a CSV format with the columns PathologyID, PathologyName, and Word. Use 'DEFINITE' as the "Word" if the pathology is explicitly mentioned as confirmed present, and 'POSSIBLE' if the pathology is explicitly suggested as a possibility. If a pathology is explicitly excluded, list it as 'ABSENT'. Please explain your answers. BEGIN RADIOLOGY IMPRESSION {IMPRESSION} END RADIOLOGY IMPRESSION List of Pathologies: {LIST OF PATHOLOGIES}.”
\end{quote}

\subsection{Fine-tune prompt for MSK impressions}\label{sec:appendixA.2}
\begin{quote}
    "Fine-tune Prompt: You are a musculoskeletal radiologist. Your task is to list only those pathologic conditions from the list of pathologies that are explicitly mentioned in the radiology impression as either possible or definite. The pathology should be included strictly if it is specifically named in the report. Do not infer the presence or absence of a pathology that is not explicitly named in the report. If a pathology is not clearly mentioned, it should not be included in the output. If a pathology is specifically excluded with phrases such as "no fracture", or "without evidence of", or "no evidence of", or "without", or "within normal limits", please do not list that pathology. Present your answer in a comma-separated list of pathology names. Include the pathology name in the output list if it is explicitly mentioned as confirmed present, or if it is explicitly suggested as a possibility. If a pathology is explicitly excluded, exclude it from the output list. BEGIN RADIOLOGY IMPRESSION {IMPRESSION} END RADIOLOGY IMPRESSION.  Here is a List of Pathologies: {lis\_of\_pathologies}."
\end{quote}

\section{Appendix B: List of pathologies}\label{sec:appendixB}
\begin{longtable}{|p{0.4\textwidth}|p{0.5\textwidth}|}
\hline
Pathological Focus & Pathologies \\ 
\hline
Tendon Pathologies & achilles tendon tear, biceps tear, complete rotator cuff tear, partial rotator cuff tear, patellar tendon tear, quadriceps tendon tear, triceps tendon tear, medial gastrocnemius tear/strain, posterior tibial tendon (ptt) tear, peroneus brevis split tear.\\ 

\hline
Ligament Pathologies & acl reconstruction with cyclops lesion, anterior cruciate ligament (acl) tear, medial collateral ligament (mcl) tear, posterior cruciate ligament (pcl) tear, meniscal root tear, radial meniscal tear, lateral meniscal tear, discoid meniscus, triangular fibrocartilage complex (tfcc) tear.\\ 

\hline

Joint and Dislocation Pathologies & acromioclavicular separation, anterior glenohumeral dislocation, posterior glenohumeral dislocation, patellar dislocation, hip dysplasia, femoroacetabular impingement, labrum tear. \\ 

\hline

Fractures and Bone Pathologies & bisphosphonate fracture, bucket handle meniscal tear, burst fracture, calcaneus fracture, compression fracture, distal radius fracture, fracture nonunion, greater tuberosity fracture, intertrochanteric fracture, intra-articular fracture, lisfranc fracture, maisonneuve fracture, mallet finger, pathologic fracture, pelvic ring fracture, scaphoid fracture with avn, segond fracture, stress fracture, subchondral insufficiency fracture, trimalleolar fracture. \\ 

\hline

Degenerative and Inflammatory Pathologies & adhesive capsulitis, ankylosing spondylitis, calcific tendinitis, calcium pyrophosphate deposition disease (cppd), carpal tunnel syndrome, de quervain's tenosynovitis, diffuse idiopathic skeletal hyperostosis, gout, hypertrophic osteoarthropathy (hpo), iliotibial band syndrome, rheumatoid arthritis, psoriatic arthritis, lupus, dermatomyositis, scleroderma, melorheostosis, myositis ossificans, pigmented villonodular synovitis, renal osteodystrophy \\ 

\hline

Infectious Pathologies & cellulitis, discitis, intraosseous abscess, necrotizing fasciitis, osteomyelitis, septic arthritis, soft tissue abscess, pott's disease \\ 

\hline
Neoplastic Pathologies (Tumors) & brown tumor, chondrosarcoma, enchondroma, fibrous dysplasia, glomus tumor, hemangioma, lipoma, liposarcoma, multiple myeloma, non-ossifying fibroma, osteochondroma, osteoid osteoma, osteosarcoma, parosteal osteosarcoma, peripheral nerve sheath tumor, synovial sarcoma \\ 

\hline
Neurological and Nerve Pathologies & baxter's neuropathy, cubital tunnel syndrome, parsonage-turner syndrome, morton's neuroma, neuropathic (charcot) arthropathy, stener lesion \\ 

\hline
Post-Surgical and Prosthetic Pathologies & arthroplasty with osteolysis or loosening, dislocated total hip arthroplasty, girdlestone procedure, periprosthetic fracture, polyethylene wear \\ 

\hline
Congenital and Developmental Pathologies & bipartite patella, calcaneonavicular coalition, lunotriquetral coalition, madelung deformity, talocalcaneal coalition \\ 

\hline
Soft Tissue and Other Pathologies & baker's cyst, biceps subluxation, elastofibroma dorsi, epicondylitis, freiberg infraction, gamekeeper's thumb, haglund's syndrome, ischiofemoral impingement, morel-lavallee, osteochondral injury, plantar fasciitis, plantar plate tear, posterior longitudinal ligament ossification, scapholunate advanced collapse, synovial osteochondromatosis \\ 

\hline
\end{longtable}

\section{Appendix C: Fine-tuned results across modalities}\label{sec:appendixC}
\begin{table*}[!h]
\centering
\small
\begin{tabular}{ccccccc}
\hline

\multicolumn{1}{c}{Model} & \multicolumn{1}{c}{Modality} & \multicolumn{1}{c}{Support} & \multicolumn{1}{c}{Precision} & \multicolumn{1}{c}{Recall} & \multicolumn{1}{c}{F1 Micro} & \multicolumn{1}{c}{F1 Macro}  \\ 

\hline
\multirow{3}{*}{GPT-4} & CR & 363 & \textbf{0.973} & 0.903 & 0.937 & 0.565  \\ 
                       & CT & 104 & 0.936 & 0.712 & 0.808 & 0.285  \\
                       & MR & 504 & 0.926 & 0.892 & 0.909 & 0.592  \\
\hline

\multirow{3}{*}{BERT} & CR & 363 & 0.93 & 0.878 & 0.903 & 0.549  \\ 
                       & CT & 104 & 0.91 & 0.81 & 0.857 & 0.344  \\
                       & MR & 504 & 0.925 & 0.875 & 0.899 & 0.65  \\
\hline

\multirow{3}{*}{Llama-3-8B} & CR & 363 & 0.929 & 0.913 & 0.921 & 0.584  \\ 
                       & CT & 104 & \textbf{0.954} & \textbf{0.89} & \textbf{0.921} & \textbf{0.399}  \\
                       & MR & 504 & 0.936 & \textbf{0.917} & 0.926 & \textbf{0.76}  \\
\hline

\multirow{3}{*}{Llama-3-8B-instruct} & CR & 363 & 0.934 & 0.911 & 0.923 & 0.588  \\ 
                       & CT & 104 & 0.929 & 0.877 & 0.902 & 0.388  \\
                       & MR & 504 & 0.924 & 0.912 & 0.918 & 0.727  \\
\hline

\multirow{3}{*}{Mistral-v0.3} & CR & 363 & 0.948 & \textbf{0.94} & \textbf{0.944} & \textbf{0.594}  \\ 
                       & CT & 104 & 0.936 & 0.902 & 0.919 & 0.409  \\
                       & MR & 504 & 0.945 & 0.932 & 0.938 & 0.751  \\
\hline
\multirow{3}{*}{Mistral-v0.3-instruct} & CR & 363 & 0.942 & 0.901 & 0.921 & 0.584  \\ 
                       & CT & 104 & 0.954 & 0.883 & 0.917 & 0.397  \\
                       & MR & 504 & \textbf{0.949} & 0.917 & \textbf{0.933}	& 0.747  \\
\hline

\end{tabular}
\label{tab:Modality}
\caption{Fine-tuned results across modalities}
\end{table*}




\end{document}